# SurgiSAM2: Fine-tuning a foundational model for surgical video anatomy segmentation and detection

**Author List:** Devanish N. Kamtam[1],* Joseph B. Shrager[1,2],* Satya Deepya Malla[1], Xiaohan Wang[6], Nicole Lin[1], Juan J. Cardona[3], Serena Yeung-Levy[5,6], Clarence Hu[4, †]

* - co-first authors, † - Senior author.

**Affiliations:**

1. Division of Thoracic Surgery, Department of Cardiothoracic Surgery, Stanford University School of Medicine, Stanford, California, USA.
2. Veterans Affairs Palo Alto Health Care System, Palo Alto, CA, USA.
3. Department of Neurosurgery, Stanford University School of Medicine, Stanford, California, USA.
4. Hotpot.ai, Palo Alto, California, USA.
5. Department of Biomedical Data Science, Stanford University, Stanford, California, USA.
6. Department of Computer Science, Stanford University, Stanford, California, USA.

**Corresponding author:**

Devanish N. Kamtam

Division of Thoracic Surgery,

Department of Cardiothoracic Surgery

Stanford University School of Medicine

300 Pasteur Drive, MC 5407

Stanford, CA 94305-5407

Email: devanish@stanford.edu

**Abstract: (300 words)**

**Background:**

The foundational segmentation models, Segmenting Anything Model (SAM) and SAM 2, have transformed segmentation by enabling remarkable zero-shot performance across diverse domains. In this study, we evaluate SAM 2 for surgical scene understanding by examining its semantic segmentation capabilities for organs/tissues both in zero-shot scenarios and after fine-tuning.

**Methods:**

We utilized five public datasets to evaluate and fine-tune SAM 2 for segmenting anatomical tissues in surgical videos/images. Fine-tuning was applied to the image encoder and mask decoder. We limited training subsets from 50 to 400 samples per class to better model real-world constraints with data acquisition. The impact of dataset size on fine-tuning performance was evaluated with weighted mean Dice coefficient (WMDC), and the results were also compared against previously reported state-of-the-art (SOTA) results.

**Results:**

SurgiSAM 2, a fine-tuned SAM 2 model, demonstrated significant improvements in segmentation performance, achieving a 17.9% relative WMDC gain compared to the baseline SAM 2. Increasing prompt points from 1 to 10 and training data scale from 50/class to 400/class enhanced performance; the best WMDC of 0.92 on the validation subset was achieved with 10 prompt points and 400 samples per class. On the test subset, this model outperformed prior SOTA methods in 24/30 (80%) of the classes with a WMDC of 0.91 using 10-point prompts. Notably, SurgiSAM 2 generalized effectively to unseen organ classes, achieving SOTA on 7/9 (77.8%) of them. Heavily dissected tissues and similar appearing organs such as small and large intestines remained challenging.

**Conclusion:**

SAM 2 achieves remarkable zero-shot and fine-tuned performance for surgical scene segmentation, surpassing prior SOTA models across several organ classes of diverse datasets. This suggests immense potential for enabling automated/semi-automated annotation pipelines, thereby decreasing the burden of annotations facilitating several surgical applications.

## Introduction:

Foundational models have transformed the field of natural language processing [1,2]. These models, pre-trained on massive datasets in a task-agnostic manner, can be fine-tuned for downstream tasks that differ from their initial training objectives. Their ability to generalize and deliver remarkable zero-shot performance on novel tasks, i.e. perform tasks without any prior task-specific training, offers substantial advantages in reducing the need for expensive dataset creation and curation. Computer vision has benefited by leveraging large general-purpose models for generative purposes[3] and achieving state-of-the-art (SOTA) performance on other vision tasks such as image classification and object detection[4–6]. Recently, with the release of Segment Anything Model (SAM)[7] and SAM 2[8] by Meta, this approach has been applied to semantic segmentation, shifting away from the traditional method of developing task-specific models[9]. This enables generalization to unseen datasets and tasks for various biomedical and clinical applications with minimal adaptation.

In biomedicine, semantic segmentation is indispensable, particularly in medical imaging for disease diagnosis, treatment planning, and disease monitoring. However, as noted, the field is currently dominated by inflexible, task-specific models. The zero-shot performance of SAM and SAM 2 on medical images has been modest[10], with inconsistent results across datasets and tasks. Low contrast, indistinct borders, small, or amorphous objects [11–13] and other complexities of medical imaging contribute to these limitations. However, SAM and SAM 2 have demonstrated the potential to surpass SOTA performance in tasks involving large, well-defined objects [14,15]. Moreover, several studies leveraging large-scale medical datasets have shown significant performance improvements by fine-tuning various components of SAM – image encoder and mask decoder in Biomedical SAM 2[16], mask decoder only in MedSAM[17], low-rank-adaptation (LoRA) fine-tuning in SAMed[18], and customized adapter modules in Medical SAM Adapter[19]. This yielded impressive outcomes, often matching or surpassing SOTA fully supervised task-specific models.

While SAM and SAM 2 have primarily been evaluated for applications in computer-aided diagnosis, semantic segmentation may serve another critical medical application – surgical scene understanding. Achieving pixel-perfect identification of structures is essential for accurately interpreting surgical scenes. Success with this could greatly advance the future of surgery and surgical education, enabling precise spatio-temporal tracking of tools, tissues, and their interactions that can facilitate downstream applications such as real-time surgical navigation, automated skill assessment, and even, ultimately, autonomous robotic surgery.

Tool segmentation has been relatively straightforward due to tools' distinct boundaries and striking contrast against background tissues [20,21]. However, the zero-shot performance of SAM 2 on surgical segmentation tasks involving live tissues in a surgical context remains unexplored. Furthermore, the limited representation of surgical data in the training sources of both SAM and SAM 2 presents an opportunity to significantly improve performance by fine-tuning SAM 2 on surgical video data. Given the cumbersome and costly nature of obtaining labeled training images in this domain, we utilized public surgical segmentation datasets to comprehensively evaluate SAM 2's capabilities.

The contributions of this paper are as follows:

We evaluated zero-shot promptable segmentation capabilities of SAM 2 for anatomical tissues in surgical videos.

- We fine-tuned SAM 2 on public surgical segmentation datasets, achieving SOTA performance across multiple organ classes included in the fine-tuning process, while also demonstrating generalization to several unseen organ classes that were not part of the fine-tuning process.
- We evaluated the impact of constrained datasets on fine-tuning performance to mimic real-world challenges associated with obtaining labeled surgical video datasets for segmentation. We demonstrate that substantial fine-tuning performance gains can be achieved with as few as 50 images per class.
- A generalized foundation model can greatly aid surgical scene understanding, where labeled datasets are scarce, and segmentation is challenging due to the need for time-consuming and expensive annotations. This fine-tuned model could significantly improve annotation efficiency in creating surgical video datasets and thereby facilitate adoption of computer vision models for various biomedical and clinical applications.

## Methods:

### Preliminary SAM 2 architecture:

The SAM 2 architecture is a versatile segmentation model designed for both image and video segmentation tasks. It builds upon SAM by integrating advanced capabilities for temporal image processing. The core components of the SAM 2 architecture include an image encoder, prompt encoder, and mask decoder, with novel components such as memory encoder, memory attention mechanism, and memory bank to enhance segmentation performance in videos (Figure 1). While SAM 2 allows modeling of temporal context across frames in a video, we utilized it exclusively for the segmentation of individual images in this study.

**Figure 1:**

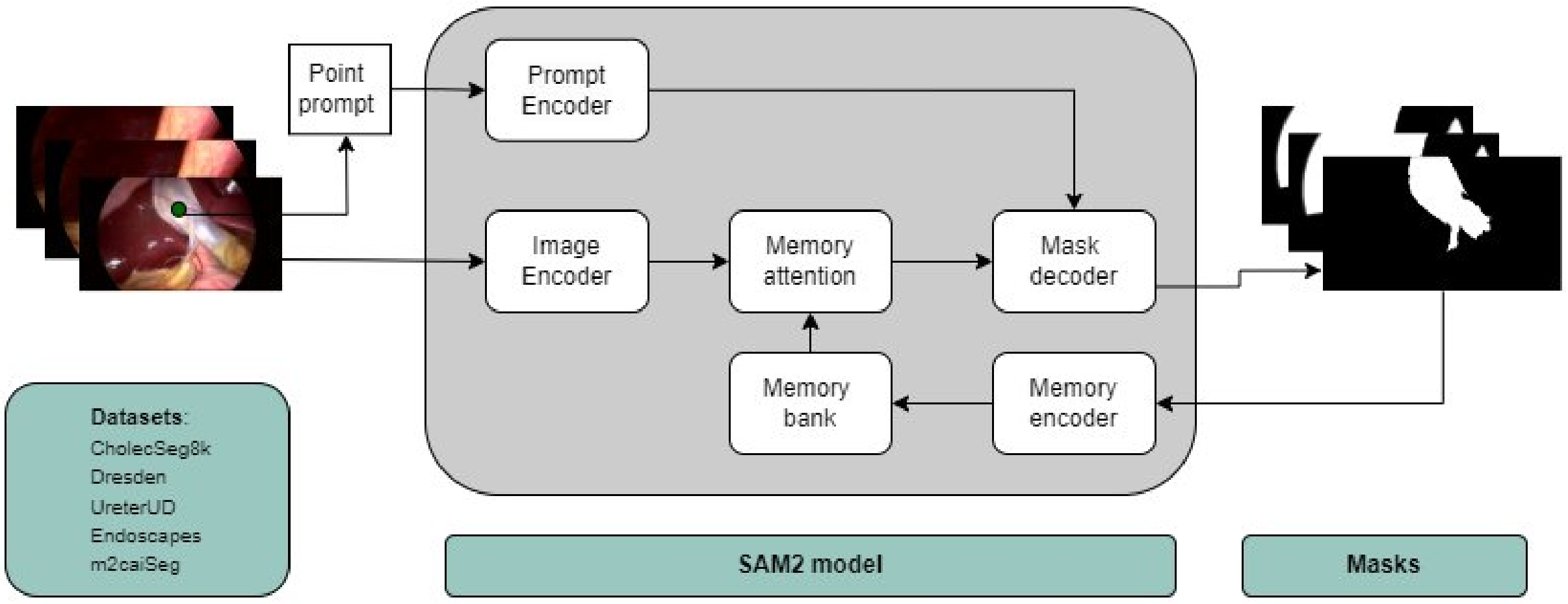

### Dataset characteristics:

To comprehensively assess the segmentation performance of SAM 2 over surgical scenes, we utilized five surgical video datasets with varying numbers of annotations for several organ/tissue classes from different surgical specialties. These datasets were chosen based on a literature review of prior studies with publicly available surgical video datasets. These datasets are as follows:

1. CholecSeg8k[22] comprises 8,080 unique laparoscopic images focused on cholecystectomy procedures. It includes 12 segmentation classes - abdominal wall, blood, connective tissue, cystic duct, fat, gallbladder, gastrointestinal tract, grasper, hepatic vein, L-hook electrocautery, liver, and liver ligament.
2. Dresden[23] comprises 2,431 unique laparoscopic images focused on colorectal procedures. It includes 11 classes - abdominal wall, colon, inferior mesenteric artery, intestinal veins, liver, pancreas, small intestine, spleen, stomach, ureter, and vesicular glands.
3. UreterUD[24] comprises 586 unique laparoscopic images focused on urological procedures. It includes 3 classes - ureter, uterine artery, and nerves.
4. Endoscapes[25] comprises 493 unique laparoscopic images focused on cholecystectomy. It includes 6 classes - cystic duct, cystic artery, cystic plate, gallbladder, hepatocystic triangle, and instruments.

5. m2caiSeg[26] comprises 299 unique images from minimally invasive abdominal surgeries. It includes 17 classes - artery, bile, bipolar, blood, clipper, fat, gallbladder, grasper, hook, intestine, irrigator, liver, scissors, specimen-bag, trocar, unknown, and upper wall.

These datasets represent a diverse set of challenges in terms of class complexity, number of classes, and surgical context, providing a comprehensive evaluation framework for fine-tuning SAM 2 for surgical scene segmentation tasks and determining its capabilities for surgical scene understanding. Instrument classes, when present in the dataset, were segmented but were excluded from the weighted mean Dice coefficient (WMDC). These results are reported in Supplementary file 1.

**Dataset preprocessing and splitting**:

The images/frames in the datasets were utilized in their original form without any pre-processing. The multi-class masks were split into individual class-wise masks for all datasets. The dataset preprocessing scripts/notebooks are provided in the Github repository (https://github.com/Devanish31/SurgiSAM2).

All datasets were split into train, validation, and test subsets, ensuring patient-wise splitting across all classes within each dataset. The splits were as follows: CholecSeg8k [13/2/2 (patients)], Dresden [90/5/5 (%)], Endoscapes [201/41/40(patients)], UreterUD [70/15/15 (%)], and m2caiSeg [80/10/10 (%)]. Additional quality-control was performed in the m2caiSeg dataset to remove poor-quality masks with issues such as empty masks, mismatch between mask and image sizes, and masks with <50 pixels mask area.

**Evaluation pipeline and training data**

We randomly extracted points from the ground truth masks to mimic the user interactively providing points for prompting SAM 2 for segmentation. Points were sampled incrementally from 1 to 10, and the zero-shot segmentation performance on the validation subsets was evaluated in intervals of 2. We did not explore additional prompt variations, such as combining positive and negative prompts or incorporating bounding box prompts, as the primary focus of this study was not on prompt engineering but rather to evaluate model performance. Nonetheless, these other approaches could potentially enhance performance.

The training subsets were used for fine-tuning the SAM 2 model, with segmentation performance tracked on the validation subsets using 10-point prompts. To investigate the impact of data scale, fine-tuning was performed using varying amounts of data per class (50, 100, 200, and 400 samples) from the training subset, assessing whether increased data volume improves performance. We deliberately restricted the dataset to fewer than 400 samples for two reasons: (1) the primary goal was to investigate SAM 2 under real-world conditions where surgical training data is typically scarce; and (2) limiting to 400 samples per class ensured a more balanced representation across all classes because several categories lacked sufficient masks at data scales of 200 and 400 avoiding a larger class imbalance beyond 400 per class.

A total of 21 unique organ/tissue classes were selected from the overall 30 organ/tissue classes across the datasets and chosen for fine-tuning. The segmentation performance with 10-point prompts was compared against the baseline SAM 2 model's performance under similar conditions.

The best-performing fine-tuned model checkpoint across all data scales was evaluated on the test subset. Its performance was compared across various classes and tasks to that of other algorithms or model architectures reported in the original dataset papers.

**Segmentation performance metrics, analyses, and visualization**

Segmentation quality was determined using several standard metrics to evaluate the overlap/ agreement between predicted and ground truth masks. The following metrics were computed:

1. **Intersection over Union (IoU):** IoU measures the overlap between the predicted mask and the ground truth mask, calculated as the ratio of their intersection to their union.

   Specifically, it is defined as:

$$\text{IoU} = \begin{cases} \frac{\text{Intersection}}{\text{Union}}, & \text{if Union} > 0 \\ 0, & \text{otherwise.} \end{cases}$$

   IoU provides a robust measure of agreement by penalizing both false positives and false negatives.

2. **Dice coefficient:** The Dice coefficient evaluates the similarity between the predicted and ground truth masks, calculated as:

   The Dice coefficient is defined as:

$$\text{Dice} = \begin{cases} \frac{2\cdot\text{Intersection}}{\text{Predicted Sum}+\text{Ground Truth Sum}}, & \text{if Predicted Sum} + \text{Ground Truth Sum} > 0 \\ 0, & \text{otherwise.} \end{cases}$$

   This metric emphasizes overlap by weighting the intersection relative to the total size of the predicted and ground truth masks.

3. **Precision:** Precision quantifies the proportion of correctly predicted pixels among all pixels in the predicted mask [True positive / (True positive + False positive)], defined as:

   It is defined as:

$$\text{Precision} = \begin{cases} \frac{\text{Intersection}}{\text{Predicted Sum}}, & \text{if Predicted Sum} > 0 \\ 0, & \text{otherwise.} \end{cases}$$

   High precision indicates fewer false positives.

4. **Recall:** Recall measures the proportion of correctly predicted pixels out of all pixels in the ground truth mask [True positive / (True positive + False negative)], calculated as:

   It is defined as:

$$\text{Recall} = \begin{cases} \frac{\text{Intersection}}{\text{Ground Truth Sum}}, & \text{if Ground Truth Sum} > 0 \\ 0, & \text{otherwise.} \end{cases}$$

High recall indicates fewer false negatives.

These metrics were computed for all examples/masks across the validation and test subsets and averaged for each class. While all metrics were calculated, the most commonly used segmentation metric, Dice coefficient is presented in the main text for brevity. Additionally, a weighted mean of the Dice coefficient was calculated as the WMDC with the results of tissue classes of all datasets, with weights determined by the number of examples in each class.

The average Dice coefficient for each class ($\text{Dice}_i$) is computed as:

$$\text{Dice}_i = \frac{1}{n_i} \sum_{j=1}^{n_i} \text{Dice}_{i,j}$$

Where:

$\text{Dice}_{i,j}$: Dice coefficient for the $j$-th example in class $i$.

$n_i$: Total number of examples in class $i$.

The overall WMDC ($\text{Dice}_{\text{weighted}}$) is then calculated as:

$$\text{Dice}_{\text{weighted}} = \frac{\sum_{i=1}^{C} n_i \cdot \text{Dice}_i}{N}$$

Where:

$N$: Total number of examples across all classes, defined as $N = \sum_{i=1}^{C} n_i$.

$C$: Total number of classes.

$n_i \cdot \text{Dice}_i$: Weighted contribution of class $i$ to the overall Dice coefficient, based on the number of examples in that class.

**Fine-tuning details and hyperparameters**

We followed the fine-tuning specifications outlined in the SAM 2 Github repository (https://github.com/facebookresearch/sam2). The SAM 2 model was fine-tuned using the AdamW optimizer with a base learning rate of $5.0\times10^{-6}$ and a cosine scheduler. The vision specific learning rate was $3.0\text{x}10^{-6}$. The weight decay was set at 0.1. The loss function incorporated mask loss, Dice loss, IoU loss, and class loss with weights of 20, 1, 1, and 1, respectively. This was consistent with the original SAM/SAM 2 training approach. Training was conducted for 30 epochs with a batch size of 1, employing data augmentation techniques such as horizontal flips, affine transformations, resizing, and color jittering. Fine-tuning was conducted using the pre-trained SAM 2.1 checkpoint (SAM 2-Hiera-B+), optimizing only the image encoder and mask decoder. All experiments were performed on a single NVIDIA A100 GPU, saving checkpoints every two epochs. Fine-tuning was carried out across varying training data scales (50, 100, 200, and 400 samples per class) including 21 unique organ/tissue classes, completing each scale within 3, 6, 11, and 16 hours, respectively. Checkpoints were saved every fifth epoch for each training scale, and the WMDC across all tissue classes was analyzed to identify the best checkpoint for each scale.

**Assessing generalizability of fine-tuned model**

To preserve SAM 2's generalized segmentation capabilities and mitigate catastrophic forgetting, we employed a low learning rate and limited the number of training epochs. We performed multi-dataset training to improve generalization across datasets and minimize overfitting to a single dataset. To determine generalizability, we assessed the segmentation performance of the best fine-tuned model checkpoint on an unseen test subset of all datasets that were split patient-wise for all classes across all datasets. We also assessed the fine-tuned model’s performance on unseen/untrained classes of the datasets (majorly m2caiSeg).

Additionally, we also evaluated the segmentation performance of SurgiSAM 2 against another biomedical foundational segmentation model, MedSAM.

**Preliminary evaluation of tissue tracking in videos: Baseline SAM 2 vs. SurgiSAM 2**

We conducted an evaluation of tissue tracking performance for both the baseline and fine-tuned SAM 2 models using surgical videos from four distinct procedures. These videos, sourced from YouTube under a Creative Commons license, featured tissues and organs not included in the training dataset: the lung in lobectomy, the ovary in hysterectomy, the appendix in appendectomy, and the spleen in renal cyst enucleation. From each video, we extracted 7–8 short segments of 0.5 seconds in duration, resulting in approximately 100 frames per tissue class and a total of ~400 frames. To simulate a manual segmentation workflow, the first frame of each segment was prompted with 1 to 10 points, and the resulting mask generated by the video predictor model was propagated across the remaining frames. The Dice coefficient was then calculated for each organ class to compare the tracking performance of the baseline and fine-tuned models.

## Results:

### Dataset characteristics

The dataset characteristics of all included datasets are presented in Table 1 and Figure 2. The five datasets, CholecSeg8k, Dresden, Endoscapes, UreterUD, and m2caiSeg include 45,635 annotated masks across 12 segmentation classes, 13,138 annotated masks across 11 segmentation classes, 648 annotated masks across 3 segmentation classes, 1,911 masks across 6 segmentation classes, 2,044 masks across 17 segmentation classes, respectively.

**Figure 2:**

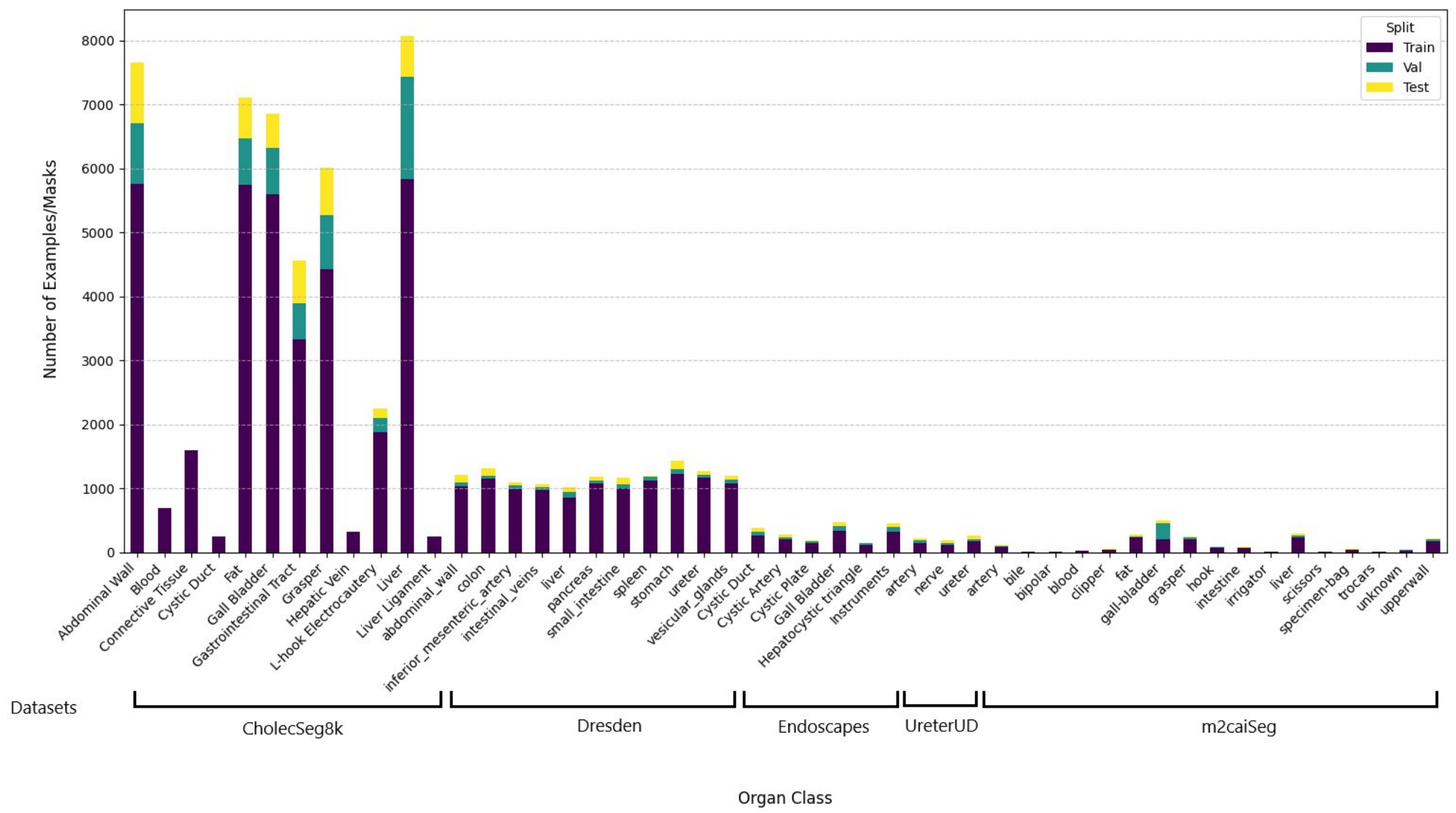

### Zero-shot evaluation: Impact of model backbone and prompt quantity

The performance of baseline SAM 2 was evaluated using Hiera Large and Hiera Base Plus backbones with 1 to 10 prompt points. The WMDC improved progressively with increasing numbers of prompts, reaching the highest performance at 10 points for both models (Figure 3). Hiera Large consistently outperformed Hiera Base Plus across most classes, with overall WMDC of 0.84 and 0.78, respectively with 10 prompt points (Figure 3). The largest performance gains were observed in structures such as cystic artery (+0.36), cystic plate (+0.74), and the liver (+0.27), and vesicular glands (+0.19). Following this, segmentation was only performed with the Hiera Base Plus model for computational purposes.

**Figure 3:**

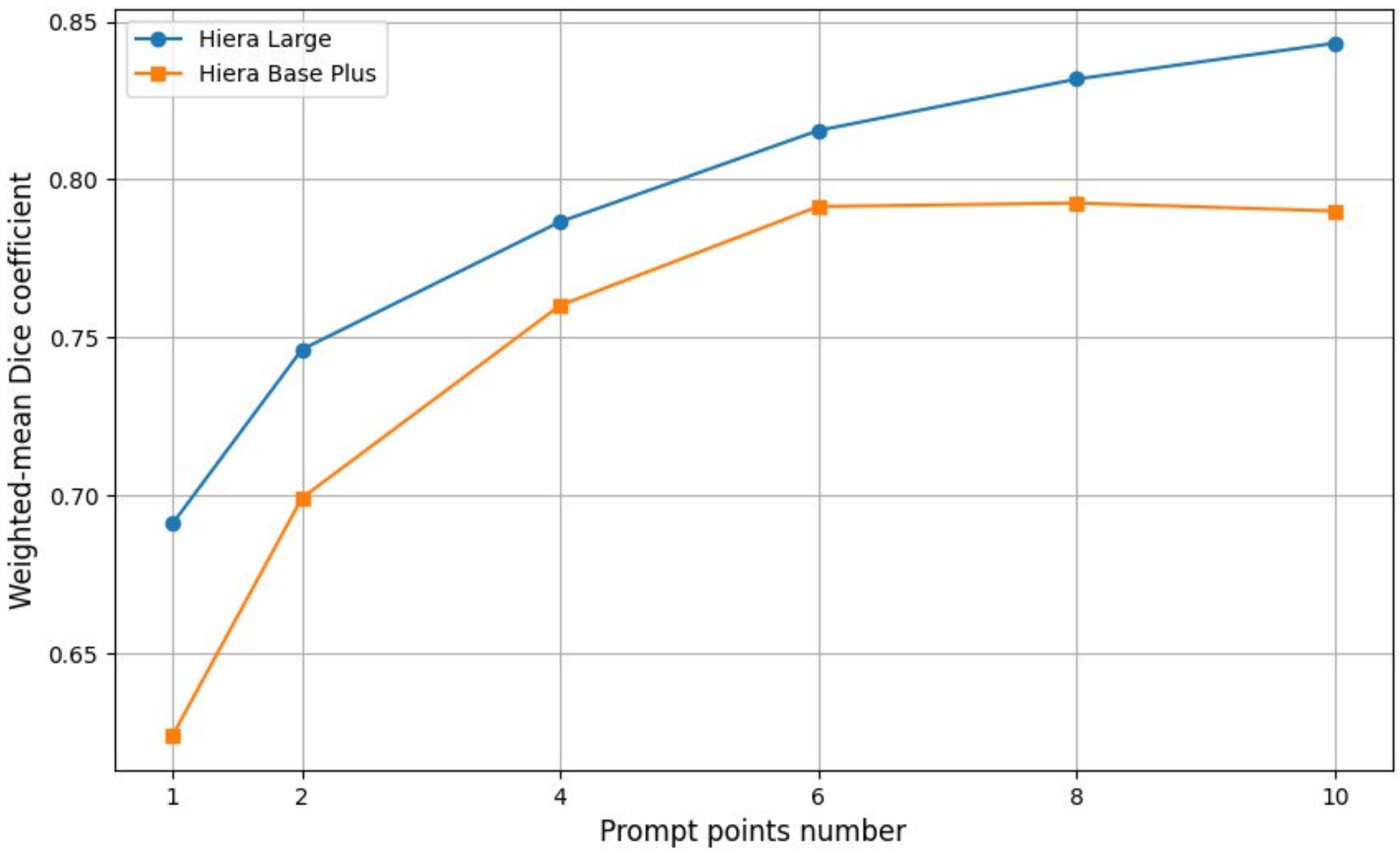

## Fine-tuning performance across training data scales

Fine-tuning SAM 2 (SurgiSAM 2) resulted in an absolute WMDC improvement of 0.14, representing a 17.9% relative improvement over the baseline SAM 2 Base Plus model across all tissue classes. The most substantial performance gains occurred within the first six epochs across all data scales, after which improvements were marginal (Figure 4). The effect of training data scale on model performance was assessed by fine-tuning SAM 2 with different sample sizes per class (50, 100, 200, and 400). For each data scale, the best-performing model checkpoint was identified based on the WMDC at 5-epoch intervals and used for further comparison. Notably, for all data scales (50 to 400 samples per class), performance improved only marginally beyond the 6th epoch (Supplementary file 2).

**Figure 4:**

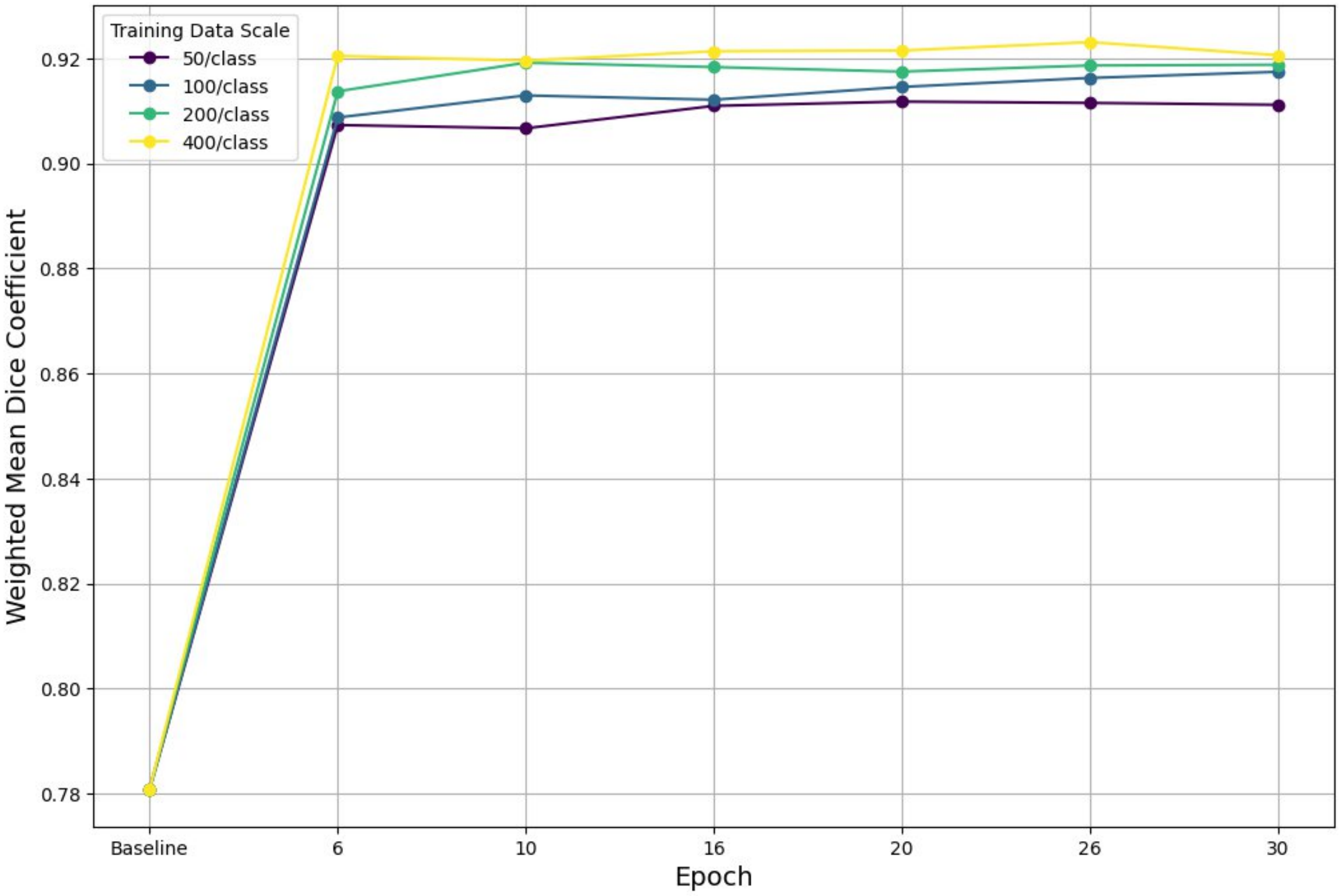

Regarding training data scale, increasing the number of samples per class beyond 50 resulted in minimal enhancement in segmentation performance. The highest performance was recorded with 400 samples per class and 10 prompts, achieving a WMDC of 0.92 (Figure 4). The benefits of data scaling were consistent across various organs and structures. Scaling the training data improved WMDC across tissue classes; however, the WMDC reduced when also considering instrument classes, given the model shift towards specifically segmenting tissues/organs (Figure 5).

**Figure 5:**

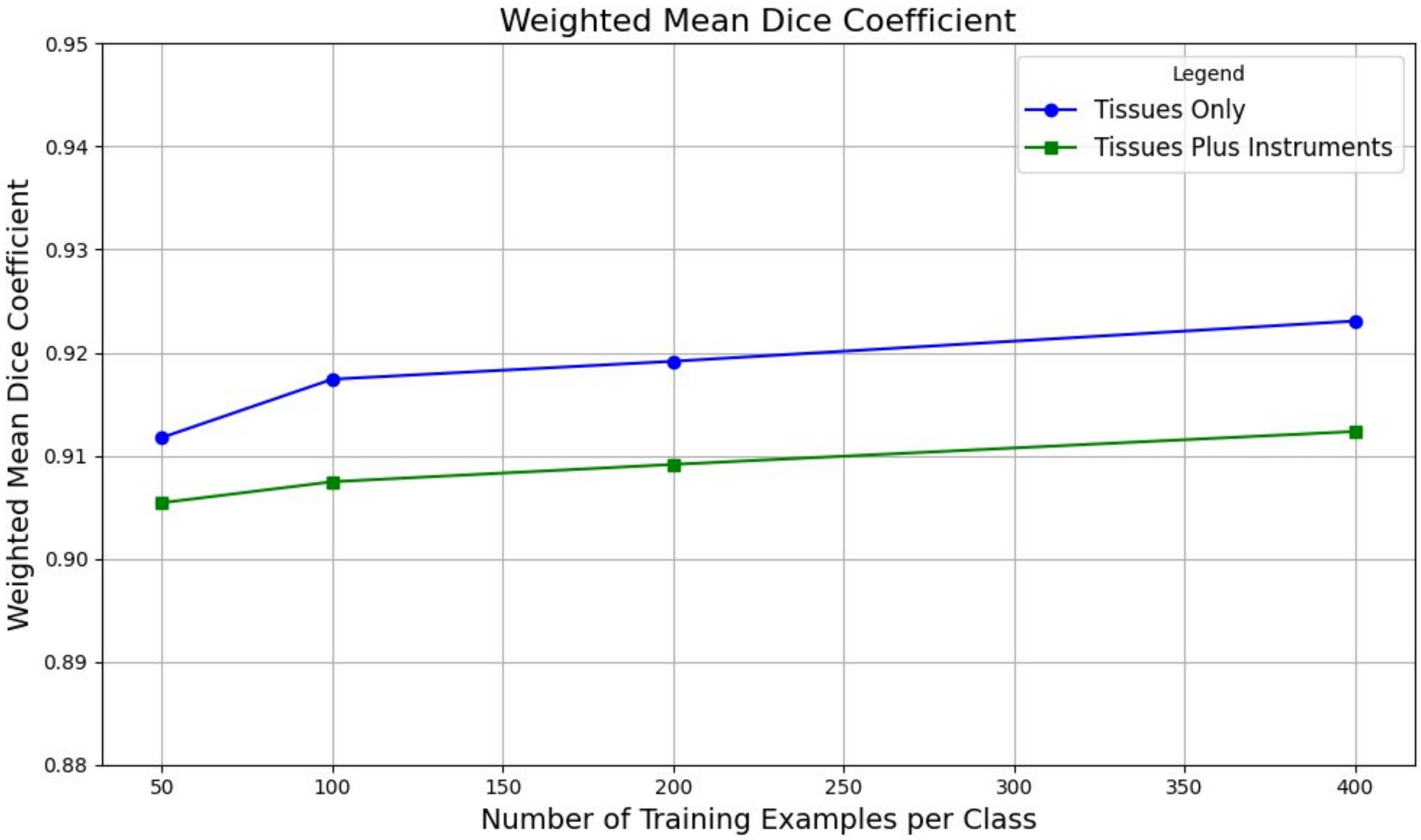

## Comparison against prior SOTA and other SAM models

SurgiSAM2's performance was evaluated against prior SOTA methods using both 1-point and 10-point prompts on test subsets. Fine-tuned SAM 2 (SurgiSAM 2) achieved substantial improvements in segmentation accuracy and consistently outperformed prior SOTA methods with 10-point prompts (24/30 classes, 80%) and even 1-point prompt (20/30 classes, 66.6%) (Table 3). Interestingly, SurgiSAM2 also outperformed the medical segmentation-specific model, MedSAM, over all organ classes (Figure 6). SurgiSAM 2 excelled in segmenting smaller and more challenging structures, with the most significant gain in organs such as the inferior mesenteric artery, cystic duct, cystic artery, uterine artery, and vesicular glands, with average Dice coefficient improvements of 0.43, 0.37, 0.32, 0.29, and 0.28, respectively. However, certain classes, including gallbladder, abdominal wall, and liver, showed only marginal improvements due to their impressive baseline performance. These findings highlight SAM's capability in handling complex segmentation tasks, especially smaller and anatomically intricate structures, reinforcing its potential as a reliable and adaptable model for surgical scene understanding.

**Figure 6:**

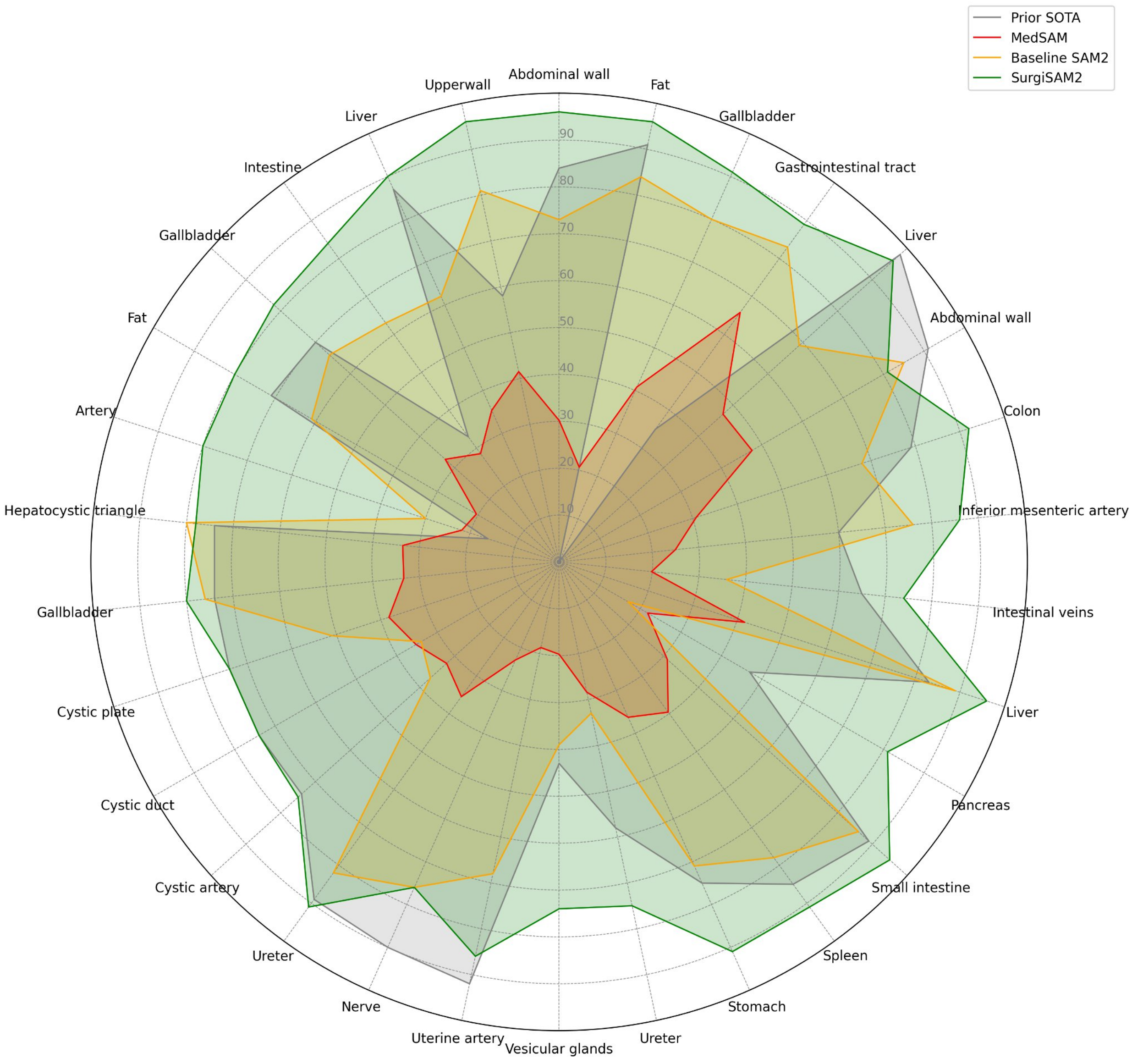

## Generalization capability

A strict, patient-wise split for training, validation, and testing was implemented to prevent data leakage. No appreciable decline in performance was observed in the test subset relative to the training subset, further supporting the model's capacity to generalize effectively to unseen examples and patients. Fine-tuning SAM 2 on 21 selected tissue classes out of the available 30, also resulted in improved Dice scores across the remaining 9 unseen classes, with an average increase of 0.17 compared to the baseline SAM 2. Notably, it achieved SOTA performance on 7 of these 9 unseen classes (77.8%). The unseen classes primarily consisted of organ-redundant categories across different datasets of the training data. This highlights the model's capacity to generalize to similar classes (organs/tissues) beyond the training datasets, demonstrating strong cross-dataset transfer.

### Qualitative assessment of successful and edge cases

To clearly demonstrate the segmentation efficacy, we performed visualization experiments showcasing examples of both the best and worst segmented cases, as determined by Dice metric, across various classes and datasets. The results are presented in Figure 7.

**Figure 7:**

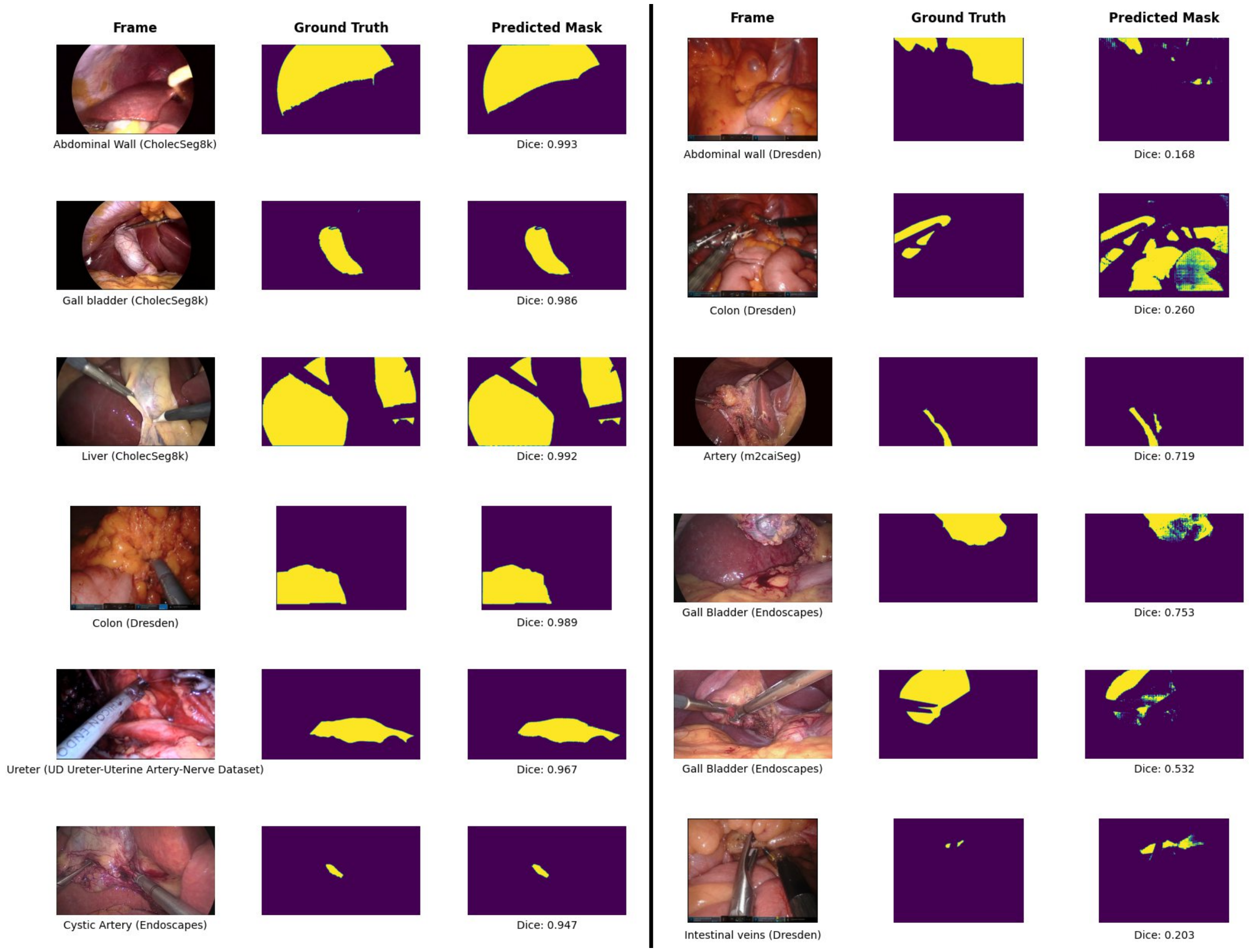

### Preliminary evaluation of tissue tracking in videos: Baseline SAM 2 vs. SurgiSAM 2

SurgiSAM 2 demonstrated a modest improvement over the baseline SAM 2 model (Figure 8) across all classes when using between 1-10 prompt points, while performing comparably at certain prompt point counts. Notably, the baseline model itself showed robust performance, with results progressively improving with increasing number of prompt points. The superior baseline performance on these videos, compared to other image datasets, can be attributed to several factors. First, the use of masks as prompts provides a dense representation of organ classes. Second, the short duration of the videos (0.5 seconds) results in minimal frame-to-frame variation. Lastly, the organs evaluated—characterized by their sharp borders—are inherently easier to segment. This experiment served as a preliminary assessment of the models' tissue tracking capabilities, which have the potential to significantly enhance annotation workflows, rather than a comprehensive evaluation of the baseline and fine-tuned SAM 2 models.

**Figure 8**:

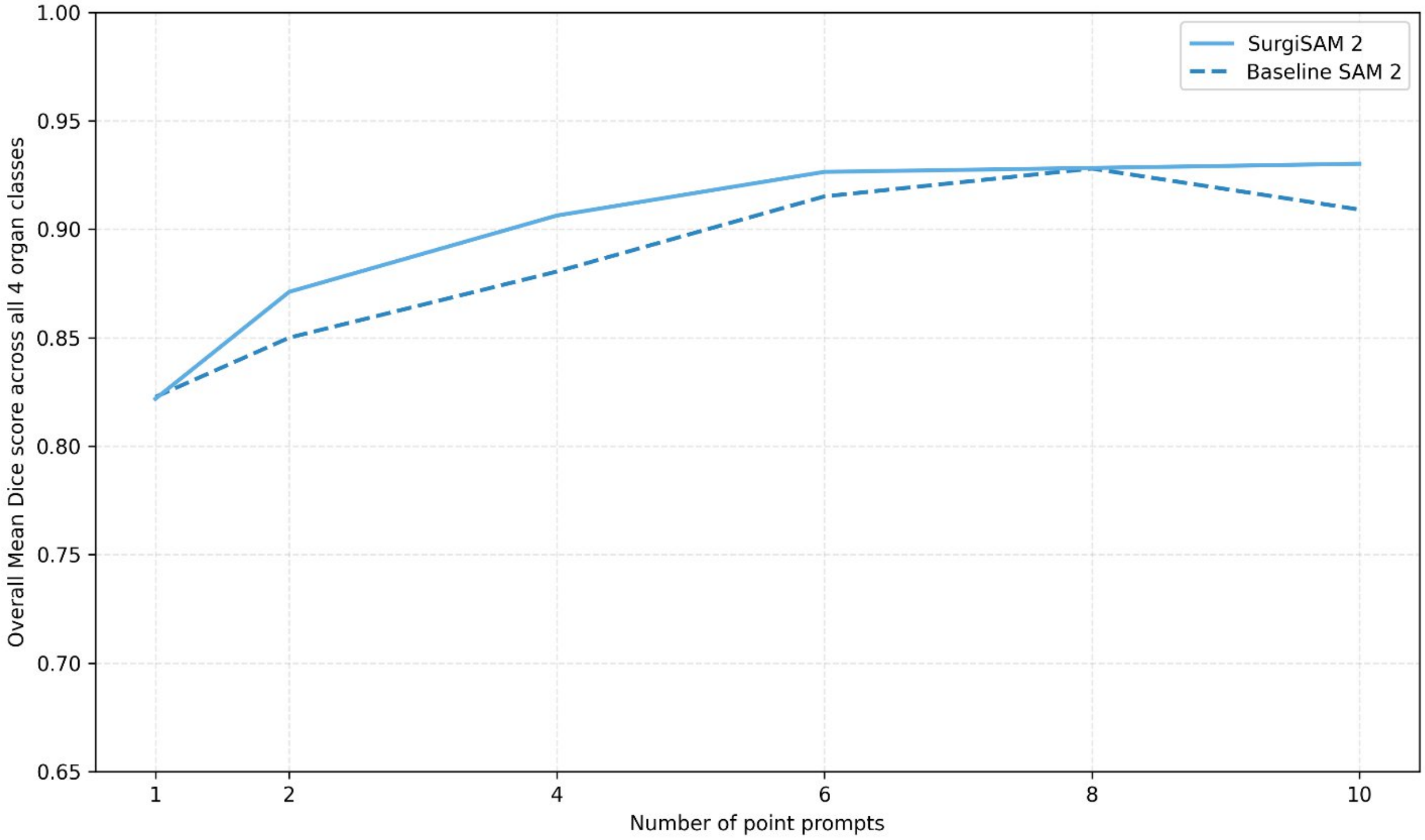

**Discussion:**

This study demonstrates the adaptability of foundational models like SAM 2 to specialized surgical data, achieving SOTA segmentation performance across a wide range of organs and tissues. Fine-tuning SAM 2 on surgical datasets delivered remarkable accuracy and generalization, often surpassing task-specific models, while requiring significantly fewer labeled samples. Unlike traditional task-specific models that demand extensive labeled datasets for each task/application, SurgiSAM 2 can leverage its generalized segmentation capabilities to provide a scalable, resource-efficient solution for fully automated or semi-automated segmentation pipelines for surgical applications. This is particularly important given how labor intensive it is to annotate surgical videos, the need for busy, trained surgeons to do this annotation, and the resulting scarcity of annotated videos.

Our findings align with recent advancements in medical image segmentation with SAM, where fine-tuning foundational models have shown significant promise, particularly when both the image encoder and the mask decoder are fine-tuned. While this study did not directly compare the fine-tuning of individual components of SAM 2, prior research suggests that models fine-tuned solely on the mask decoder deliver inferior performance[16]. Our fine-tuning approach showed a 17.9% relative improvement over baseline SAM 2 and consistently outperformed prior SOTA models across most tissue classes, without significant performance degradation on the instrument classes. Although using bounding boxes or learned feature vectors as prompts could further enhance performance [27–29], this work focused on evaluating SurgiSAM 2's potential for surgical scene segmentation rather than optimizing prompt engineering.

Remarkably, these results were achieved with only 50–400 labeled samples per class. Even with just 50 samples per class, fine-tuning resulted in only marginal performance degradation compared to using 400 samples per class. This represents a reduction of one to two orders of magnitude in training data requirements compared to fully supervised task-specific models like nnU-Net[30], while delivering comparable or slightly superior performance. Furthermore, SurgiSAM 2 excelled at generalizing to unseen tasks and datasets, achieving SOTA performance on 77.8% of unseen redundant classes from other datasets. This underscores its ability to effectively capture both low-level features and high-level organ-specific semantic features across diverse anatomical structures. The model's capacity to generalize across datasets without retraining is particularly advantageous in the surgical setting, where video annotations are costly, time-consuming, and access to highly skilled annotators is limited.

Another critical finding is SurgiSAM 2's ability to effectively segment discontinuous or amorphous structures (as shown with liver in Figure 6) that are common in surgical data but remain difficult to segment accurately. This was also observable with a single point prompt. Despite all these strengths and notable performance improvements with challenging structures such as the gastrointestinal tract, ureter, and vesicular glands, certain challenges remained. SurgiSAM 2 performed poorly on small organs that were heavily dissected, such as the cystic plate, cystic artery, cystic duct, and intestinal veins in Figure 6. Tissue dissection can unpredictably alter the appearance and feature representation of organs, making it challenging for models to accurately identify them. The presence of adipose tissue overlying abdominal organs can also complicate annotation and hinder the model's performance, as it may be challenging to distinguish and label pixels accurately as either the organ or adipose tissue. Additionally, the small intestine was sometimes misclassified as the colon when both appeared in the same frame, likely due to their similar visual characteristics or due to the class, gastrointestinal tract, in CholecSeg8k dataset having examples from both small and large intestine. Furthermore, the segmentation of abstract and dynamic anatomical concepts, such as the hepatocystic triangle, remained a significant challenge. It may be that some of the tissues that could be of greatest benefit to surgeons (and their patients) to be able to identify using segmentation models such as the common bile duct during a cholecystectomy and a ureter during a colon resection, may prove in many cases to be the most difficult to identify using segmentation models. Finally, other challenges include poor illumination and smoke from surgical cauterization that affect image quality and model performance.

Further enhancements to SurgiSAM 2's performance could be achieved by incorporating the larger image encoder, such as Hiera Large, and exploring prompting strategies such as: combination of positive and negative prompts, avoiding border regions for sampling prompts points, bounding boxes, and high dimensional learned organ-specific feature vectors [31,32]. Moreover, adopting multi-frame segmentation approaches to leverage temporal information in videos, rather than relying solely on isolated images as in this study, could further enhance segmentation. Memory optimization methods, such as efficient frame pruning[33], could further facilitate surgical video segmentation in intraoperative settings.

The ability of SurgiSAM 2 to generalize and perform well even with limited training data has significant clinical implications. By enabling segmentation in scenarios where annotated datasets are scarce, SurgiSAM 2 addresses a critical bottleneck in the adoption of CV models for surgical applications. Currently, the rate-limiting step for integrating AI into surgical workflows is the need for extensive manual annotation of training data, which is both time-consuming and resource-intensive. With SurgiSAM 2's remarkable segmentation capabilities, surgeons can leverage AI to enhance surgical scene understanding, even in underrepresented or novel surgical scenarios. This reduces the dependency on large-scale annotation efforts, paving the way for broader clinical adoption of artificial intelligence-driven tools in surgical applications such as real-time intraoperative guidance, automated skill assessment, and robotic surgery.

The limitations of this study are herein acknowledged. While this study utilized all public datasets with segmentation masks for tissues, these likely do not comprehensively represent the full diversity of organs and structures encountered during surgical procedures. This may limit the generalizability of SurgiSAM2 to the more broader range of procedures and anatomical structures encountered in surgical settings. Moreover, the random convenient sampling of frames from the original datasets for the range of training data scales may overrepresent scenes from certain aspects of surgery. Incorporating a more representative dataset through manual curation by surgeons, or automated sampling of images based on methods such as cosine similarity for clustering, could further improve performance. Finally, the potential for achieving higher performance with the full dataset scale was not evaluated due to computational constraints.

**Conclusion:**

In conclusion, SAM 2 demonstrates remarkable zero-shot performance and exhibits significant improvements with fine-tuning across multiple organ classes from diverse datasets. The fine-tuned SAM 2, SurgiSAM 2 underscores the potential of foundational segmentation models to offer robust, generalizable, and cost-effective solutions for surgical scene segmentation, even with limited training data. SurgiSAM 2 paves the way for scalable segmentation solutions by enabling semi-automatic pipelines that significantly reduce manual annotation requirements. This holds immense potential for surgical scene understanding, by facilitating accurate spatio-temporal tracking of tissues and instruments, thereby enabling clinical applications such as real-time surgical navigation, automated skill assessment and autonomous robotic surgery.

**Code and Data Availability Statement:** The underlying code [and training/validation/test splits of all datasets] for this study and data with the final fine-tuned checkpoint are available on Github [https://github.com/Devanish31/SurgiSAM2] and Figshare [10.6084/m9.figshare.28489961], respectively.

**CRediT authorship contribution statement:** DNK – Conceptualization, Data curation, Formal analysis, Methodology, Supervision, Writing – original draft, Writing – review and editing; JBS - Conceptualization, Methodology, Supervision, Writing – review and editing; SDM - Conceptualization, Data curation, Writing – original draft, Writing – review and editing; XW- Conceptualization, Methodology, Supervision, Writing – review and editing; NL - Conceptualization, Methodology, Writing – review and editing; JJC – Conceptualization, Data curation, Methodology, Writing – review and editing; SY - Conceptualization, Methodology, Supervision, Writing – review and editing; CH - Conceptualization, Methodology, Supervision, Writing – review and editing. All authors read and approved of the final manuscript.

**Declaration of generative AI and AI-assisted technologies:** During the preparation of this work, the author(s) utilized ChatGPT to assist with rephrasing and refining the writing in the manuscript. After using this tool/service, the author(s) reviewed and edited the content as needed and take(s) full responsibility for the content of the published article.

**Conflicts of interest**: Joseph Shrager: Consulting - Becton Dickinson; Lungpacer; Serena Yeung - Research collaborations with Intuitive Surgical Inc. and Surgical Safety Technologies; Clarence Hu: Founder – Hotpot.ai; Other authors have nothing to declare.

**Funding**: None

**Acknowledgements:** None

**Tables:**

**Table 1:** Characteristics of publicly- available datasets used for fine-tuning and evaluation of the SAM 2 model, including dataset source, number of unique images, segmentation classes (test split), and annotated masks.

| Dataset paper author (citation) | Abbreviated dataset name | Number of unique images | Number of classes (in test split) | Number of masks |
|---|---|---|---|---|
| **W.Y. Hong et al.[21]** | CholecSeg8k | 8,080 | 12 (7)* | 45,635 |
| **Mathias Carstens et al. [22]** | Dresden | 2,431 | 11 (11) | 13,138 |
| **Norbert Serban et al.[23]** | UreterUD | 586 | 3 (3) | 648 |
| **A Murali et al. [24]** | Endoscapes | 493 | 6 (6) | 1,911 |
| **Salman Maqbool et al.[25]** | m2caiSeg | 299 | 17 (17) | 2,044 |

*Certain classes were excluded from validation and test subsets due to the inability to split the dataset in a patient-wise manner, which was necessary to maintain generalizability.

**Table 2**: Dice coefficients for segmentation performance of Hiera Large and Hiera Base Plus backbones across multiple datasets and anatomical structures, evaluated with 1 to 10 prompt points. The weighted mean Dice coefficient for all classes is also provided to summarize overall performance trends. The best Dice score for each class for each model is highlighted in bold.

| **Datasets and their organ classes (n=number of examples in validation subset)** | **Hiera-Large** | | | | | | **Hiera-Base-Plus** | | | | | |
|---|---|---|---|---|---|---|---|---|---|---|---|---|
| | **Number of point prompts** | | | | | | **Number of point prompts** | | | | | |
| | **1** | **2** | **4** | **6** | **8** | **10** | **1** | **2** | **4** | **6** | **8** | **10** |
| **CholecSeg8k (4548)** | | | | | | | | | | | | |
| Abdominal wall (951) | 0.89 | 0.93 | 0.95 | 0.95 | **0.96** | 0.95 | 0.78 | 0.83 | 0.90 | 0.93 | **0.94** | **0.94** |
| Fat (720) | 0.70 | 0.77 | 0.76 | 0.77 | 0.79 | **0.80** | 0.72 | 0.76 | 0.77 | 0.82 | 0.84 | **0.85** |
| Gallbladder (720) | 0.84 | 0.86 | 0.88 | **0.89** | **0.89** | 0.88 | 0.78 | 0.84 | 0.88 | **0.88** | **0.88** | 0.87 |
| Gastrointestinal tract (557) | 0.83 | 0.82 | 0.91 | 0.91 | 0.92 | **0.93** | 0.60 | 0.68 | **0.76** | **0.76** | 0.72 | 0.69 |
| Liver (1600) | 0.56 | 0.68 | 0.73 | 0.77 | 0.80 | **0.83** | 0.53 | 0.64 | 0.72 | 0.76 | **0.77** | 0.75 |
| **Dresden (621)** | | | | | | | | | | | | |
| Abdominal wall (56) | 0.71 | 0.78 | 0.85 | 0.88 | 0.91 | **0.92** | 0.67 | 0.71 | 0.77 | 0.79 | **0.80** | 0.79 |
| Colon (52) | 0.67 | 0.70 | 0.76 | 0.82 | 0.82 | **0.85** | 0.65 | 0.66 | 0.76 | **0.77** | 0.76 | 0.71 |
| Inferior mesenteric artery (61) | 0.37 | 0.40 | 0.48 | 0.54 | 0.56 | **0.57** | 0.30 | 0.41 | **0.47** | 0.46 | 0.45 | 0.36 |
| Intestinal veins (49) | 0.50 | 0.55 | 0.63 | 0.67 | 0.67 | **0.68** | 0.29 | 0.45 | 0.58 | 0.59 | 0.56 | 0.51 |
| Liver (83) | 0.79 | 0.83 | 0.87 | **0.89** | **0.89** | **0.89** | 0.76 | 0.86 | 0.87 | **0.88** | **0.88** | **0.88** |
| Pancreas (42) | 0.75 | 0.72 | 0.79 | 0.82 | **0.83** | **0.83** | 0.61 | 0.68 | **0.73** | 0.72 | 0.70 | 0.65 |
| Small intestine (53) | 0.67 | 0.82 | 0.87 | **0.89** | 0.88 | **0.89** | 0.74 | 0.85 | 0.90 | 0.90 | 0.90 | **0.91** |
| Spleen (50) | 0.82 | 0.85 | 0.86 | 0.90 | **0.95** | 0.93 | 0.81 | 0.85 | 0.88 | **0.90** | **0.90** | 0.88 |
| Stomach (78) | 0.79 | 0.81 | 0.85 | 0.89 | **0.91** | **0.91** | 0.64 | 0.72 | 0.82 | 0.81 | **0.82** | 0.78 |
| Ureter (43) | 0.28 | 0.32 | 0.45 | 0.48 | **0.50** | **0.50** | 0.21 | 0.26 | 0.35 | 0.36 | **0.37** | 0.28 |
| Vesicular glands (54) | 0.45 | 0.52 | 0.59 | 0.66 | **0.67** | 0.64 | 0.33 | 0.43 | 0.51 | 0.53 | **0.54** | 0.50 |
| **UreterUD (115)** | | | | | | | | | | | | |
| Uterine artery (50) | 0.56 | 0.55 | 0.64 | **0.75** | **0.75** | 0.74 | 0.52 | 0.60 | 0.68 | **0.70** | 0.65 | 0.60 |
| Nerve (30) | 0.63 | 0.67 | 0.71 | 0.76 | 0.78 | **0.79** | 0.69 | 0.68 | 0.69 | **0.72** | **0.72** | 0.71 |
| Ureter (35) | 0.84 | 0.86 | 0.85 | **0.87** | 0.85 | **0.87** | 0.71 | 0.72 | 0.72 | **0.73** | 0.64 | 0.60 |
| **Endoscapes (226)** | | | | | | | | | | | | |
| Cystic artery (34) | 0.17 | 0.19 | 0.29 | 0.43 | 0.50 | **0.53** | 0.16 | 0.23 | 0.38 | **0.43** | **0.43** | 0.40 |
| Cystic duct (65) | 0.27 | 0.31 | 0.36 | 0.46 | 0.50 | **0.53** | 0.28 | 0.28 | 0.36 | 0.43 | **0.44** | 0.41 |
| Cystic plate (29) | 0.20 | 0.17 | 0.38 | 0.56 | 0.60 | **0.67** | 0.23 | 0.36 | 0.53 | 0.67 | **0.71** | 0.62 |
| Gallbladder (73) | 0.67 | 0.68 | 0.74 | **0.84** | **0.84** | 0.83 | 0.62 | 0.71 | 0.77 | 0.79 | 0.82 | **0.83** |
| Hepatocystic triangle (25) | 0.27 | 0.35 | 0.46 | 0.45 | 0.51 | **0.53** | 0.30 | 0.44 | 0.53 | **0.63** | 0.56 | 0.52 |
| **m2caiSeg (317)** | | | | | | | | | | | | |
| Artery (11) | 0.22 | 0.23 | 0.26 | 0.28 | **0.29** | 0.22 | 0.26 | 0.30 | 0.30 | **0.37** | 0.29 | 0.31 |
| Fat (28) | 0.43 | 0.49 | 0.55 | 0.55 | **0.60** | 0.43 | 0.46 | 0.51 | 0.51 | 0.62 | 0.63 | **0.64** |
| Gallbladder (249) | 0.51 | 0.50 | 0.55 | 0.66 | **0.71** | 0.51 | 0.43 | 0.49 | 0.58 | **0.65** | **0.65** | 0.63 |
| Intestine (7) | 0.65 | 0.84 | 0.80 | 0.84 | **0.86** | 0.65 | 0.64 | 0.48 | 0.67 | **0.75** | 0.72 | 0.67 |
| Liver (29) | 0.48 | 0.58 | 0.58 | 0.60 | **0.61** | 0.48 | 0.50 | 0.53 | 0.63 | 0.67 | 0.70 | **0.68** |
| Upperwall (22) | 0.73 | 0.74 | 0.78 | **0.82** | 0.79 | 0.73 | 0.56 | 0.62 | 0.73 | **0.79** | 0.77 | 0.75 |
| | | | | | | | | | | | | |
| **Weighted mean Dice coefficient (for all tissue classes)** | 0.69 | 0.75 | 0.79 | 0.82 | 0.83 | **0.84** | 0.62 | 0.70 | 0.76 | **0.79** | **0.79** | **0.79** |
| **Mean Dice coefficient (for all tissue classes)** | 0.58 | 0.62 | 0.67 | 0.72 | 0.74 | **0.75** | 0.53 | 0.59 | 0.66 | **0.69** | **0.69** | 0.66 |

**Table 3**: Quantitative performance metrics of the fine-tuned SAM 2 model across various datasets and classes, showing Dice coefficients for different training data scales (50, 100, 200, and 400 samples per class) with 10 prompt points. Comparisons with prior SOTA methods are highlighted in the last column with the deltas in parentheses, improvements in green and declines in red.

| **Datasets and their organ classes (n=number of examples in validation, test subsets)** | **Training data scale** | | | | **Prior SOTA score** | **Prior SOTA model** | **MedSAM (vit-b)** | **Test subset with 1 prompt point)** | **Test subset with 10 prompt points (delta)** |
|---|---|---|---|---|---|---|---|---|---|
| | **50/class** | **100/class** | **200/class** | **400/class** | | | | | |
| | **Number of point prompts (10)** | | | | | | | | |
| **CholecSeg8k (4548, 3611)** | | | | | | | | | |
| Abdominal wall (951, 943) | 0.97 | 0.97 | 0.97 | 0.97 | 0.84 | DynUnet[33] | 0.30 | 0.88 | **0.96 (+0.12)** |
| Fat (720, 640) | 0.94 | 0.95 | 0.96 | 0.97 | 0.91 | U-net++[33] | 0.21 | 0.89 | **0.96 (+0.05)** |
| Gallbladder (720, 537) | 0.90 | 0.90 | 0.91 | 0.91 | | Segmenter[34] | 0.41 | 0.90 | **0.91** |
| Gastrointestinal tract (557, 668) | 0.96 | 0.95 | 0.96 | 0.96 | 0.35 | UNETR[33] | 0.66 | 0.88 | **0.89 (+0.54)** |
| Liver (1600, 640) | 0.93 | 0.93 | 0.93 | 0.94 | **0.98** | uNet/uNet++/DeepLACv3+[35] | 0.47 | 0.91 | 0.96 (-0.02) |
| **Weighted mean Dice coefficient for dataset (for tissue classes)** | | | | | | | | | 0.94 |
| **Dresden (621, 851)** | | | | | | | | | |
| Abdominal wall †(56, 114) | 0.93 | 0.92 | 0.93 | 0.90 | **0.91** | SegFormer[36] | 0.48 | 0.56 | 0.81 (-0.10) |
| Colon (52, 121) | 0.89 | 0.90 | 0.89 | 0.91 | 0.79 | DeepLACv3[36] | 0.31 | 0.83 | **0.92 (+0.13)** |
| Inferior mesenteric artery (61, 44) | 0.77 | 0.78 | 0.77 | 0.79 | 0.6 | SegFormer[36] | 0.25 | 0.77 | **0.86 (+0.26)** |
| Intestinal veins (49, 52) | 0.74 | 0.81 | 0.78 | 0.78 | 0.65 | SegFormer[36] | 0.20 | 0.72 | **0.74 (+0.09)** |
| Liver † (83, 81) | 0.93 | 0.93 | 0.93 | 0.94 | 0.83 | SegFormer[36] | 0.42 | 0.88 | **0.96 (+0.13)** |
| Pancreas (42, 51) | 0.84 | 0.87 | 0.86 | 0.88 | 0.47 | SegFormer[36] | 0.22 | 0.78 | **0.81 (+0.34)** |
| Small intestine (53, 108) | 0.95 | 0.95 | 0.95 | 0.96 | 0.89 | SegFormer[36] | 0.31 | 0.90 | **0.95 (+0.06)** |
| Spleen (50, 14) | 0.97 | 0.96 | 0.97 | 0.97 | 0.85 | SegFormer[36] | 0.40 | 0.87 | **0.91 (+0.06)** |
| Stomach (78, 129) | 0.90 | 0.91 | 0.90 | 0.93 | 0.75 | SegFormer[36] | 0.36 | 0.91 | **0.91 (+0.16)** |
| Ureter (43, 71) | 0.50 | 0.54 | 0.57 | 0.54 | 0.58 | SegFormer[36] | 0.28 | 0.68 | **0.75 (+0.17)** |
| Vesicular glands (54, 66) | 0.76 | 0.80 | 0.80 | 0.78 | 0.43 | SegFormer[36] | 0.20 | 0.63 | **0.74 (+0.31)** |
| **Weighted mean Dice coefficient for dataset (for tissue classes)** | | | | | | | | | 0.89 |
| **UreterUD (115, 114)** | | | | | | | | | |
| Uterine artery (50, 22) | 0.89 | 0.88 | 0.90 | 0.89 | **0.92** | U-Net[23] | 0.19 | 0.79 | 0.86 (-0.06) |
| Nerve (30, 38) | 0.71 | 0.72 | 0.78 | 0.78 | **0.90** | U-Net[23] | 0.23 | 0.77 | 0.76 (-0.14) |
| Ureter (35, 54) | 0.86 | 0.87 | 0.86 | 0.86 | 0.89 | U-Net[23] | 0.36 | 0.90 | **0.91 (+0.02)** |
| **Weighted mean Dice coefficient for dataset (for tissue classes)** | | | | | | | | | 0.85 |
| **Endoscapes (226, 183)** | | | | | | Only single metric reported | | | |
| Cystic artery (34, 47) | 0.66 | 0.71 | 0.73 | 0.72 | 0.74 | Mask2Former[24] | 0.32 | 0.70 | **0.75 (+0.01)** |
| Cystic duct (65, 53) | 0.72 | 0.77 | 0.75 | 0.79 | **0.74** | Mask2Former[24] | 0.35 | 0.55 | **0.74 (0.00)** |
| Cystic plate (29, 18) | 0.72 | 0.77 | 0.79 | 0.75 | **0.74** | Mask2Former[24] | 0.38 | 0.59 | **0.74 (0.00)** |
| Gallbladder † (73, 58) | 0.84 | 0.83 | 0.85 | 0.84 | 0.74 | Mask2Former[24] | 0.33 | 0.58 | **0.80 (+0.06)** |
| Hepatocystic triangle (25, 7) | 0.64 | 0.67 | 0.67 | 0.62 | 0.74 | Mask2Former[24] | 0.34 | **0.78 (+0.04)** | 0.66 |
| **Weighted mean Dice coefficient for dataset (for tissue classes)** | | | | | | | | | 0.75 |
| **m2caiSeg (317, 154)** | | | | | | | | | |
| Artery † (11, 12) | 0.71 | 0.74 | 0.72 | 0.69 | 0.16 | Custom encoder-decoder CNN[25] | 0.22 | **0.80 (+0.64)** | 0.78 (+0.62) |
| Fat † (28, 30) | 0.78 | 0.79 | 0.81 | 0.78 | 0.71 | Custom encoder-decoder CNN[25] | 0.20 | 0.61 | **0.80 (+0.09)** |
| Gallbladder † (249, 49) | 0.81 | 0.82 | 0.83 | 0.79 | 0.7 | Custom encoder-decoder CNN[25] | 0.33 | 0.55 | **0.82 (+0.12)** |
| Intestine † (7, 9) | 0.90 | 0.89 | 0.87 | 0.90 | 0.33 | Custom encoder-decoder CNN[25] | 0.29 | 0.78 | **0.84 (+0.51)** |
| Liver † (29, 31) | 0.88 | 0.87 | 0.88 | 0.90 | 0.87 | Custom encoder-decoder CNN[25] | 0.35 | 0.74 | **0.90 (+0.03)** |
| Upper wall (22, 23) | 0.96 | 0.95 | 0.96 | 0.95 | 0.58 | Custom encoder-decoder CNN[25] | 0.42 | 0.91 | **0.96 (+0.38)** |
| **Weighted mean Dice coefficient for dataset (for tissue classes)** | | | | | | | | | 0.85 |
| | | | | | | | | | |
| **Weighted mean Dice coefficient (for all tissue classes)††** | 0.91 | 0.92 | 0.92 | **0.92** | **-** | | 0.38 | 0.85 | **0.91** |
| **Mean Dice coefficient (for all tissue classes)** | 0.83 | 0.85 | 0.85 | **0.85** | **-** | | 0.33 | 0.77 | **0.85** |

***SOTA - state-of-the-art**

Highlighted in bold are the best mean Dice scores for each organ in a comparison between prior SOTA and the SurgiSAM 2 performance on the test subset.

Better than prior SOTA is presented in green and worse than prior SOTA in red. The delta in performance between test subset and prior SOTA is reported in parentheses

† - Classes excluded from fine-tuning

†† - To address the disproportionate impact of a single dataset (CholecSeg8k) on the weighted mean Dice coefficient (WMDC), we also reported the mean Dice coefficient (MDC), calculated as the unweighted average of Dice scores across all classes.

**Figures:**

**Figure 1:** Diagrammatic representation of the SAM 2 model and its components.

**Figure 2:** The number of annotated masks (x-axis) for the five datasets (CholecSeg8k, Dresden, Endoscapes, UreterUD, and m2caiSeg), categorized by training, validation, and testing splits across all organ/tissue classes (y-axis). The CholecSeg8k dataset constituted the largest portion across all dataset splits.

**Figure 3:** Performance comparison of Hiera Large and Hiera Base Plus backbones using weighted mean Dice coefficient (x-axis), evaluated with 1 to 10 prompt points (y-axis) across segmentation tasks. The performance of Hiera-Large and Hiera-Base-Plus increases with progressively increasing number of prompts plateauing near the higher number of prompts.

**Figure 4:** Weighted mean Dice coefficient improvements (y-axis) during fine-tuning of SAM 2 with different training data scales (50, 100, 200, and 400 samples per class) (x-axis) across epochs. While all the training data scales demonstrate significant improvement from the baseline, there remains a performance gap even after prolonged training between 50 and 400 examples per class. In addition, most of the performance gains of fine-tuning are achieved as early as 6 epochs.

**Figure 5:** Weighted mean Dice coefficients (y-axis) for "Tissues Classes Only" and "Tissues Plus Instruments Classes" scenarios with single point prompt and 10-point prompts and progressively increasing training data scale (x-axis). In both the "Tissues Classes Only" and the "Tissues Plus Instruments Classes", performance progressively increases with increasing scale of data, but only marginally.

* Segmentation metrics of the instrument classes are reported in the Supplementary files 1 & 2.

**Figure 6:** Multi-class segmentation performance of different models in surgical images. Comparison of prior SOTA (state-of-the-art) (grey), MedSAM (red), baseline SAM2 (orange), and SurgiSAM2 (green) across several anatomical structures. Values shown as percentages on radial axes range from 0-100% (0.0-1.0 mean Dice scores for the organ class). SurgiSAM 2 outperforms all other models across the majority of organ and tissue classes. (*MedSAM 2 was not employed in this study due to the fine-tuning being conducted on the smaller variant of SAM 2, containing 39 million parameters. Comparing this version to the base-plus model, which has 89 million parameters and was fine-tuned in this study, may not provide a fair assessment).

**Figure 7:** Qualitative assessment of SAM's segmentation performance, illustrating best-performing (left) and worst-performing (right) predictions across different datasets and anatomical structures. SurgiSAM 2 performs exceedingly well even with smaller organs (spleen, ureter), discontinuous organs (liver), and deformed organs (gallbladder when being handled). SurgiSAM 2 may struggle when smaller organs undergo heavy dissection (cystic artery, cystic duct, intestinal veins), abstract anatomical concepts like hepatocystic triangle, and similar looking organs (small intestine and large intestine).

**Figure 8.** Comparative video tracking performance of baseline SAM 2 and SurgiSAM 2 using mean Dice scores (y-axis). The dashed line represents the performance of the baseline SAM 2 model, while the solid line corresponds to the fine-tuned SurgiSAM 2 model. Both models exhibit improved segmentation accuracy with an increasing number of point prompts (x-axis). SurgiSAM 2 marginally outperformed the baseline model across most prompt counts, achieving higher Dice scores. Performance of both models plateaus around 8–10 point prompts, with SurgiSAM 2 maintaining marginally superior Dice score.